\documentclass[runningheads]{llncs}
\usepackage{graphicx}
\usepackage{comment}
\usepackage{amsmath,amssymb} 
\usepackage{color}
\usepackage{array}
\usepackage{multirow}
\usepackage{subcaption}
\usepackage{booktabs}
\usepackage{pifont}
\newcommand{\cmark}{\text{\ding{51}}}

\usepackage[hidelinks]{hyperref}

\makeatletter
\newcommand*\fsize{\dimexpr\f@size pt\relax}
\makeatother
\renewcommand{\orcidID}[1]{\href{https://orcid.org/#1}{\includegraphics[height=1\fsize]{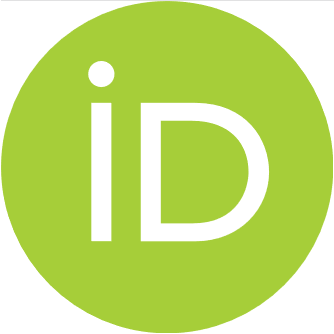}}}

\begin{document}
\pagestyle{headings}
\mainmatter
\def\CVPPPSubNumber{4}  
\def\ECCVSubNumber{\CVPPPSubNumber} 
\title{Improving Pixel Embedding Learning through Intermediate Distance Regression Supervision for Instance Segmentation} 

\titlerunning{Intermediate Distance Regression Supervision for Instance Segmentation}
\author{Yuli Wu \orcidID{0000-0002-6216-4911} \and
	Long Chen \orcidID{0000-0002-5280-4727} \and
	Dorit Merhof \orcidID{0000-0002-1672-2185}}
\authorrunning{Y. Wu et al.}
%
\institute{Institute of Imaging and Computer Vision, RWTH Aachen, Germany\\
\email{yuli.wu@rwth-aachen.de, \{long.chen, dorit.merhof\}@lfb.rwth-aachen.de}}


\maketitle

\begin{abstract}
	As a proposal-free approach, instance segmentation through pixel embedding learning and clustering is gaining more emphasis. Compared with bounding box refinement approaches, such as Mask R-CNN, it has potential advantages in handling complex shapes and dense objects. In this work, we propose a simple, yet highly effective, architecture for object-aware embedding learning. A distance regression module is incorporated into our architecture to generate seeds for fast clustering. At the same time, we show that the features learned by the distance regression module are able to promote the accuracy of learned object-aware embeddings significantly. By simply concatenating features of the distance regression module to the images as inputs of the embedding module, the mSBD scores on the CVPPP Leaf Segmentation Challenge can be further improved by more than 8\% compared to the identical set-up without concatenation, yielding the best overall result amongst the leaderboard at CodaLab.

\end{abstract}

\section{Introduction}
Instance segmentation aims to label each individual object, which is critical to many biological and medical applications, such as plant phenotyping and cell quantification. Learning object-aware pixel embeddings is one of the trends in the field of instance segmentation. The embedding is essentially a high-dimensional representation of each pixel. To achieve instance segmentation, pixel embeddings of the same object should be located relatively close in the learned embedding space, while those of different objects should be discriminable.

The loss usually consists of two terms: the between-instance loss term \(\mathcal{L}_{inter}\) and the within-instance loss term \(\mathcal{L}_{intra}\). The former term \(\mathcal{L}_{inter}\) encourages different-instance embeddings to be located far away from each other, while the latter term \(\mathcal{L}_{intra}\) encourages same-instance embeddings to stay together. Two most popular metrics used to describe the similarities of embeddings are Euclidean distance and cosine distance. Although the pixel embedding approaches have gained success in many datasets including CVPPP Leaf Segmentation Challenge~\cite{chen2019instance,de2017semantic,neven2019instance,payer2018instance}, the trained embedding space is far from optimal.

Our idea was indirectly inspired by the “easy task first” concept behind curriculum learning~\cite{bengio2019cl}. Distance regression predicts the distance from a pixel to the object boundary and is used in~\cite{chen2019instance,schmidt2018}, for example, as an auxiliary module. We have empirically found that the distance regression module is relatively easy to train on many datasets. Considering that the learned features by the distance regression module should be already recognizable for distinguishing instances, we prefix the embedding module with a distance regression module to promote the embedding learning process. 

The main contributions of this paper are summarized as follows:
\begin{enumerate}
    \item We propose an architecture to promote the pixel embedding learning by utilizing features learned from the distance regression module, which significantly improves the performance in the CVPPP Leaf Segmentation Challenge~\cite{Leaf}. Our overall mean Symmetric Best Dice (mSBD) score is at the top position of the leaderboard with 0.879 by paper submission. Furthermore, the average of mSBD scores on Arabidopsis images (testing sets A1, A2, A4) outperforms the second best results from three different teams by over 3\%, namely from 0.883 to 0.917;
    \item We conduct a number of ablation experiments in terms of the stacked U-Net architecture, different types of concatenative layers and varied loss formats, to validate our architecture and also supplement some experimental vacancies in this field. 
\end{enumerate}

\section{Related Work}
We roughly categorize some approaches of instance segmentation into two groups with respect to the overall pipeline: \textit{instance-first} approaches and \textit{one-stage} approaches. \textit{Instance-first} approaches exploit the instance-level bounding boxes from the first-stage object detector. For example, Mask R-CNN~\cite{he2017mask} uses RPN~\cite{ren2015faster}, and recent methods like BlenderMask~\cite{chen2020blendmask} and CenterMask~\cite{lee2019centermask} are based on the anchor-free detector FCOS~\cite{tian2019fcos}. Pixel-level segmentations are then produced through subjoined refinement modules. Mask R-CNN~\cite{he2017mask} constructs a lightweight segmentation network with consecutive convolutional layers, while the Blender Module and Spatial Attention-Guided Mask (SAG-Mask) are proposed in~\cite{chen2020blendmask} and~\cite{lee2019centermask}, respectively, for a more accurate segmentation.

In contrast, \textit{one-stage} approaches predict the existence (\textit{object-ness}) and mask of objects all at once. Masks are represented as polar coordinates in~\cite{schmidt2018,xie2019polarmask}. Specifically, the model regresses the distances to the boundary along a set of fixed directions at each location. To describe more complex shapes, masks are encoded with a linear projection in~\cite{zhang2020MEInst}. 
 
Furthermore, the approaches based on pixel embedding learning, which also belong to \textit{one-stage} approaches, are becoming a new trend. They share the general pipeline of \textit{embedding and clustering}. Each pixel of input images is mapped to a high-dimensional vector (embedding), in which pixels of the same object are located closely. Then, clustering in the embedding space results in the final instance segmentation. De Brabandere and Neven~\cite{de2017semantic,neven2019instance} have proposed Euclidean distance based embedding loss for instance segmentation. Payer et al.~\cite{payer2018instance} have demonstrated embedding loss which utilizes cosine similarity and recurrent stacked hourglass network~\cite{newell2016stacked}. Chen et al.~\cite{chen2019instance} have introduced a U-Net based architecture of two heads, where the embeddings are trained with cosine embedding loss and local constraints. These two heads are distance regression head and embedding head. The distance regression head aims to provide seed candidates for clustering. Our proposed method inherits the fundamental modules from this work.

For current pixel embedding based approaches, clustering is an essential step. Mean Shift~\cite{fukunaga1975estimation} and HDBSCAN~\cite{campello2015hierarchical} are used in~\cite{de2017semantic} and~\cite{payer2018instance} respectively. In~\cite{chen2019instance,neven2019instance}, threshold based clustering is used with knowledge of the learned seeds.

\begin{figure}[t]
	\centering
	\includegraphics[width=0.98\textwidth]{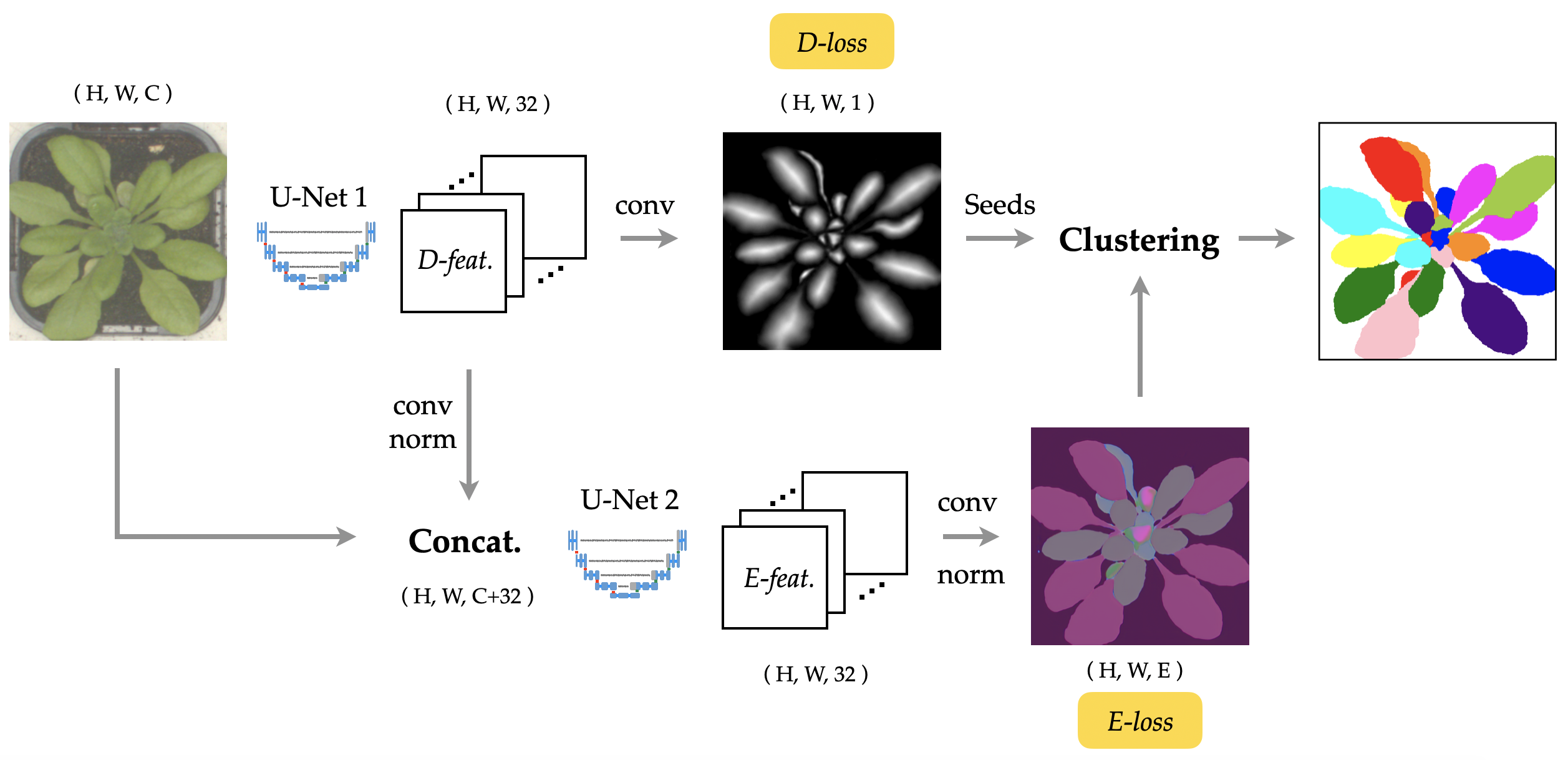}
	\caption{Processing Pipeline. Distance regression features and distmaps are learned via distance module with U-Net 1. Concatenated distance regression features and images are fed into U-Net 2, from which the embeddings are learned. Final labels are generated based on seeds (thresholded maxima of distmaps) and embeddings via angular clustering. Denotations: H, W, C, E = Dimensions of Height, Width, Channel, Embedding.}
	\label{fig:pipeline}
\end{figure}

\section{Method}
Our network consists of two cascaded parts (Fig.~\ref{fig:pipeline}): the distance regression module and the embedding module. Each module uses a U-Net architecture with a 32-dimensional output feature map as the backbone network. The learned distance and embedding feature maps are denoted as \textit{D-feat.} and \textit{E-feat.}, respectively. 

The distance regression module takes standardized images (by linearly scaling each image to have mean 0 and variance 1) as the inputs and outputs the distance map (abbreviated as \textit{distmap} in the following context) through a single convolutional layer with ReLU activation. The ground truth distmap is obtained by computing the shortest distances from pixels to the object boundary and then being normalized instance-wise with respect to the maximal value. The distance regression module is trained with Mean Squared Error (MSE) loss in this work, which is illustrated as \textit{D-loss} in Fig.~\ref{fig:pipeline}. 

Distance feature map \textit{D-feat.} learned by the distance regression module is fed to the embedding module together with the input image by concatenation. Details of the concatenation are introduced in Section~\ref{method:concatenate}. The final embeddings are obtained through a convolutional layer with linear activation, followed by L2 normalization. The embedding module is trained with the loss based on the cosine similarity and local constraints (Section~\ref{method:cosLoss}), denoted as \textit{E-loss} in Fig.~\ref{fig:pipeline}.

The embedding space trained with loss in Eq.~\ref{eq:inter} has a comprehensive geometric interpretation: embedding vectors of neighboring objects tend to be orthogonal, which simplifies the complexity of clustering. The fast \textit{angular clustering} can be effortlessly performed based on angles between embedding vectors. Firstly, seeds are obtained from distmaps by fetching local maxima with a trivial threshold (selected as 70\% of the global maximum in an image). After that, all neighboring pixels within the angular range \(\delta_a\) of a seed are collected to form a cluster. In this work, we use \(\delta_a=45 \,\mathrm{deg}\) for all experiments. At last, the labels outside of the officially provided ground truth foreground masks are omitted. 

\subsection{Cosine Embedding Loss with Local Constraints}
\label{method:cosLoss}
For the embedding module training, we build upon the loss format from~\cite{chen2019instance}. The training loss, denoted as \textit{E-loss} in Fig.~\ref{fig:pipeline}, is defined based on the cosine similarity\linebreak \(\mathcal{S}_{cos}(\mathbf{e_1}, \mathbf{e_2})  =\mathbf{e_1}^{T} \mathbf{e_2}/(\lVert \mathbf{e_1}\rVert\lVert\mathbf{e_2}\rVert)\) and is formularized as: 
\begin{equation}
	\begin{aligned}
	\mathcal{L}_{emb} &= \lambda \cdot \mathcal{L}_{inter} + \mathcal{L}_{intra} \\
	\mathcal{L}_{inter} &= \frac{1}{C} { \sum_{c_A=1}^{C} \frac{1}{|\mathbf{N}_{c_A}|} \sum_{c_B \in \mathbf{N}_{c_A}} } { \mathcal{S}_{cos}}(\boldsymbol{\mu_{c_A}} \,,\, \boldsymbol{\mu_{c_B}}) \\
	\mathcal{L}_{intra} &= \frac{1}{C} \sum^{C}_{c=1}  \frac{1}{E_c} \sum^{E_c}_{i=1} \Big[1- \mathcal{S}_{cos} (\boldsymbol{e_i} \,,\, \boldsymbol{\mu_c})\Big]\, ,\\
	\end{aligned}
\label{eq:inter}
\end{equation}
where the embedding loss is defined as the weighted sum of the between-instance loss term \(\mathcal{L}_{inter}\) and within-instance loss term \(\mathcal{L}_{intra}\) with the weighting factor \(\lambda \). \(\boldsymbol{e}\) and \(\boldsymbol{\mu}\) represents the pixel embedding vector and the mean embedding of an object, respectively. \(C\) denotes the number of objects, while the number of pixels of a single object \(c\) is denoted as \(E_c\). \(\mathbf{N}_{c_A} \) represents the set of neighboring objects around the object \(c_A\) and \(|\mathbf{N}_{c_A}|\) is the number of neighbors. 

The between-instance loss term \(\mathcal{L}_{inter}\) encourages the embeddings of different object to be separated, while the within-instance loss term \(\mathcal{L}_{intra}\) punishes the case where pixel embeddings of the same object diverge from the mean. In addition, the local constraints of this loss only force neighboring objects to form separable clusters in the embedding space. The benefits of local constraints and the comparison with the global constraint are demonstrated in Section~\ref{exp:constraint}.

\subsection{Feature Concatenative Layer}\label{method:concatenate}
The feature map \textit{D-feat.} learned by the distance regression module is firstly transformed to the desired dimensions (shown with an example of 32 in Fig.~\ref{fig:pipeline}) via a convolutional layer and then L2 normalized pixel-wise along through the feature channels before being concatenated to the images. Our experiment shows that the feature map normalization is critical to a stable training process.

\begin{figure}
	\centering
	\includegraphics[width=0.65\textwidth]{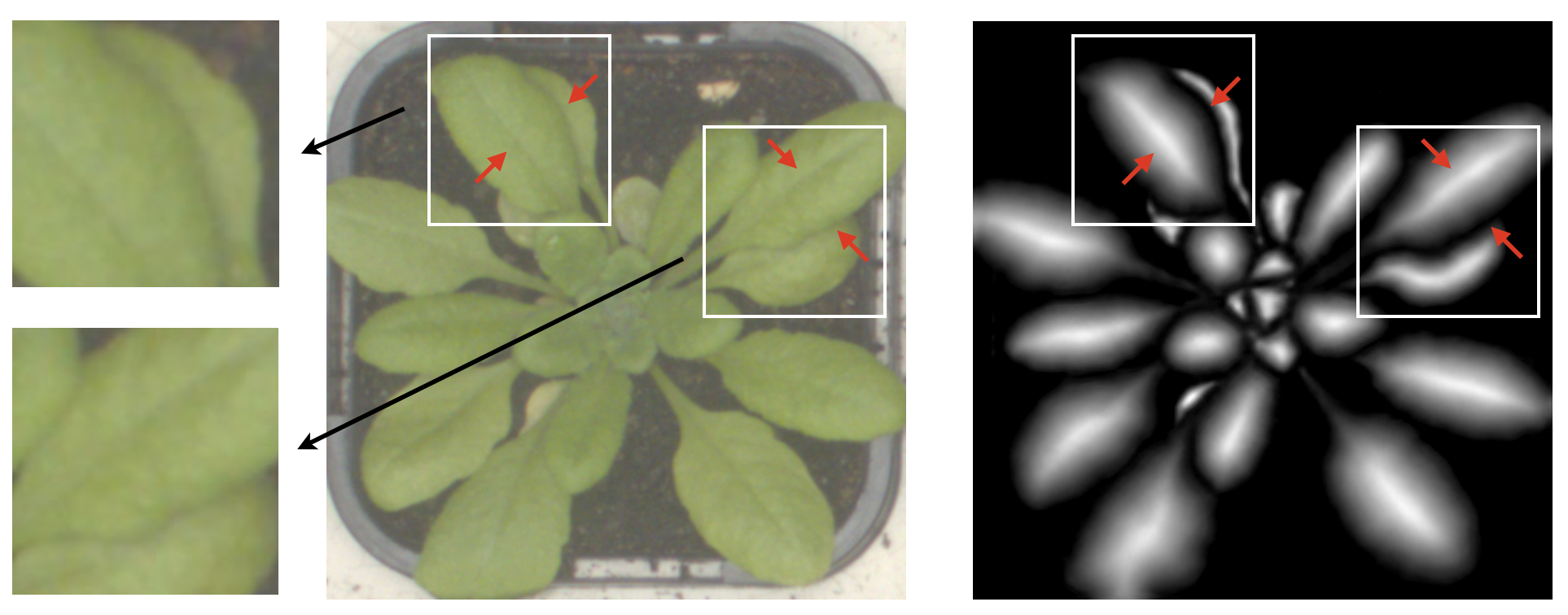}
	\caption{Ambiguity between Leaf Boundary and Leaf Midvein. Although the embedding space learned with U-Net often fails at such locations, distmap (right) is able to distinguish them well: lower values (darker areas) indicate boundaries and higher values (brighter areas) indicate midveins.}
	\label{fig:leaf_diff}
\end{figure}

As illustrated in Fig.~\ref{fig:leaf_diff}, the difference between leaf boundary and leaf midvein (primary vein) is ambiguous. The learned embeddings by the U-Net architecture~\cite{chen2019instance} often fail at those locations. However, the distmaps are able to tell the difference with lower values representing leaf boundaries and higher values representing leaf midveins. From another perspective, the distmap, which gives an approximate outline of objects, can be interpreted as a \textit{object-ness} score, the pixel-wise probability about existence of object. In addition, as proposed by~\cite{Novotny18b}, mixing convolutional operations with the pixel location helps constructing dense pixel embeddings that can separate object instances. From this perspective, the distance regression features can indirectly provide location information to the subsequent module.

To this end, we construct a two-stage architecture, as depicted in Fig.~\ref{fig:pipeline}, by forwarding the distance regression features to the embedding module. And the concatenation of the distance regression features and images can bring in best performance in the experiments. We term the distance features as concatenative layer in between the stacked U-Nets as \textit{intermediate distance regression supervision}.

In the experiments, other different features have also been tested to forward: the 1-dimensional distmap, 8-dimensional distance features, 32-dimensional distance features, 32-dimensional embedding features, concatenated 16-dimensional distance features and 16-dimensional embedding features. Inspired by~\cite{neven2019instance,Novotny18b}, we have also evaluated the performance of augmenting the input image with x- and y-coordinates.

\begin{figure}[t]
	\centering	
	\begin{subfigure}{0.8\textwidth}
		\centering
    	\includegraphics[width=\textwidth]{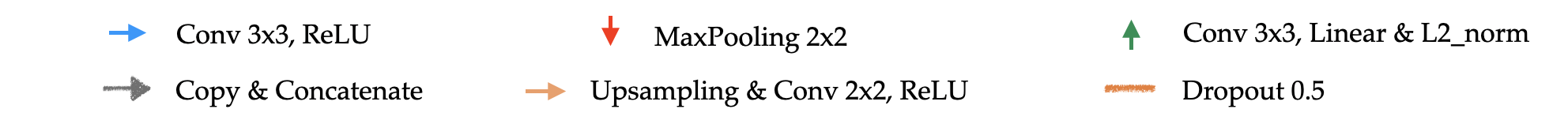}
	\end{subfigure}\\
	\begin{subfigure}{0.9\textwidth}
		\centering
    	\includegraphics[width=\textwidth]{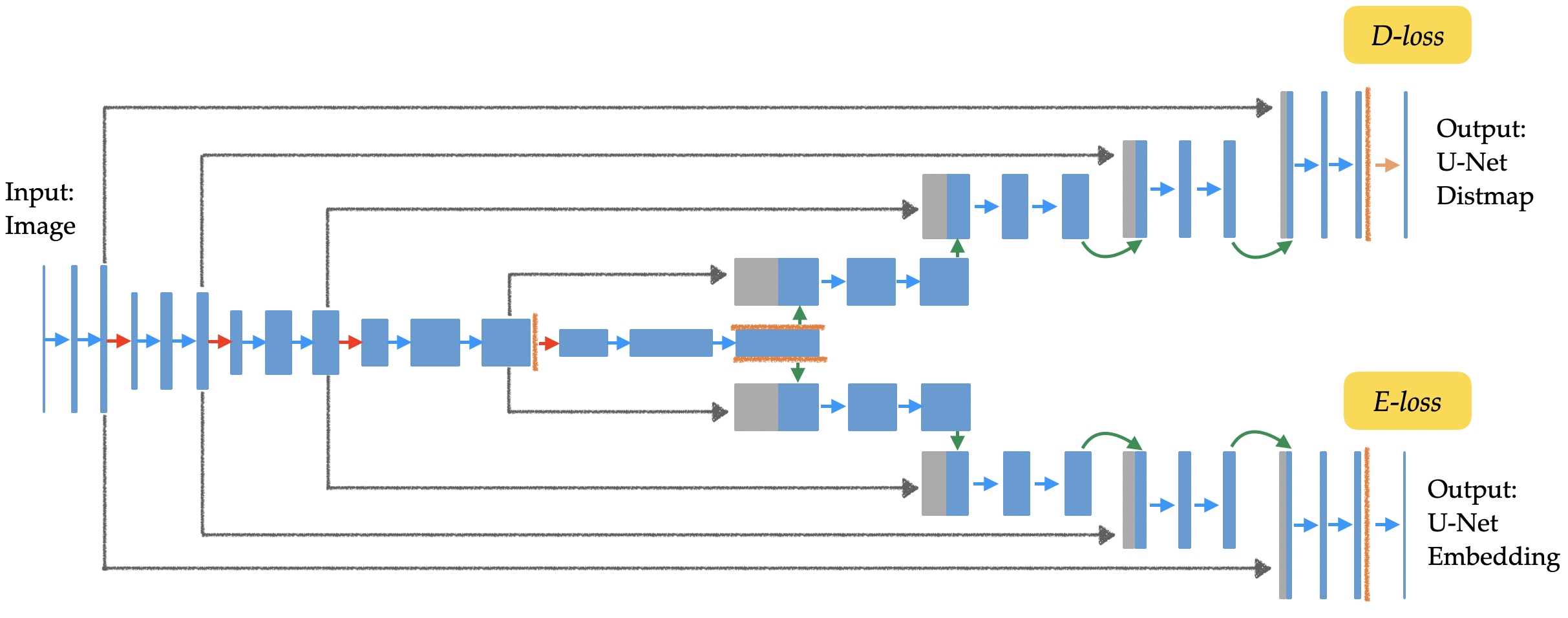}
    	\caption{U-Net with Two Heads.}
    	\label{fig:u_net}	
	\end{subfigure}\\\bigskip
	\begin{subfigure}{0.9\textwidth}
		\centering
    	\includegraphics[width=\textwidth]{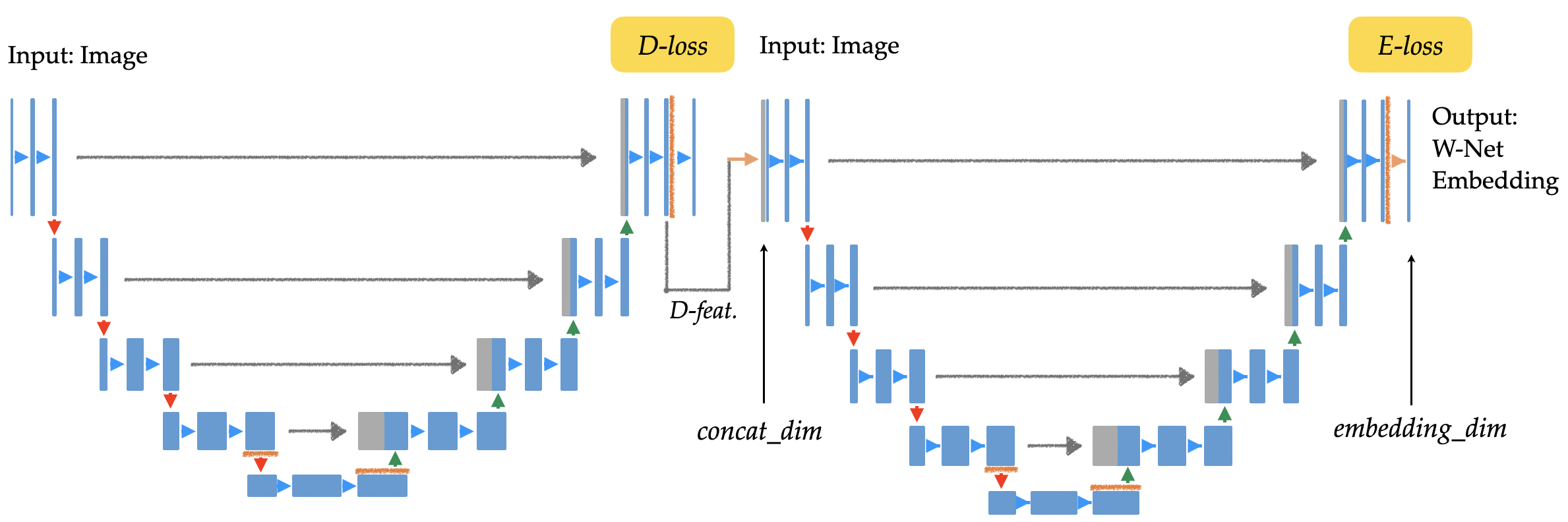}
    	\caption{W-Net with Intermediate Distance Regression Supervision.}
    	\label{fig:w_net}	
	\end{subfigure}
	\caption{Network Architectures of U-Net and W-Net.}
	\label{fig:uwnet}
\end{figure}

\subsection{From U-Net to W-Net}
We abbreviate the proposed network as W-Net to differ from the existing U-Net with two heads, although the novelty and characteristic are not fully represented: the distance regression features as intermediate supervision and the cosine embedding loss with local constraints. 

In Fig.~\ref{fig:uwnet}, the detailed architectures of U-Net with two heads and W-Net with intermediate distance regression supervision are illustrated. The parallel distance and embedding heads of U-Net (Fig.~\ref{fig:u_net}) are modified towards the serial distance and embedding modules in W-Net (Fig.~\ref{fig:w_net}). Apart from the types of concatenative layer as discussed previously, we have also investigated the final dimensions of embeddings as another hyper-parameter, denoted as \textit{embedding\_dim} in Fig.~\ref{fig:w_net}. The corresponding ablation experiments can be found in Section~\ref{ssec:dim}.

\section{Experiments}\label{sec:ex}
Ablation experiments are conducted with U-Net and W-Net, as depicted in Fig.~\ref{fig:uwnet}. The training loss is the sum of the distance regression loss (ReLU+MSE) and the cosine embedding loss with local constraints (Eq.~\ref{eq:inter}). The latest CodaLab dataset of CVPPP2017 LSC is used as training set without augmentation. Model parameters are initialized by He Normal~\cite{He_2015_ICCV} and optimized by Adam~\cite{kingma2014adam}. The initial learning rate is set to 0.0001 and scheduled with exponential decay, with the decay period being set to 5000 steps and the decay rate 0.9. The batch size is set to 4 in most experiments, or 2 if high embedding dimensions are used. The maximal training epochs are set to 500. We show mSBD scores of testing set from CodaLab as the evaluation metric.

\begin{figure}
	\centering	
	\begin{subfigure}{0.48\textwidth}
		\centering
    	\includegraphics[width=\textwidth]{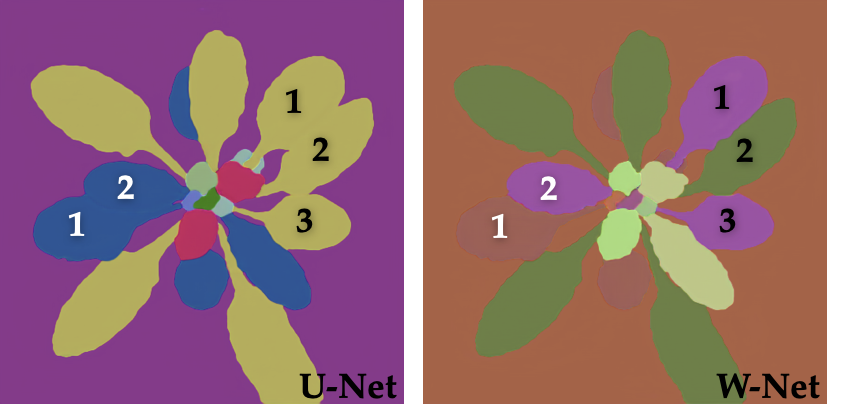}
	\end{subfigure}
	\quad
	\begin{subfigure}{0.48\textwidth}
		\centering
    	\includegraphics[width=\textwidth]{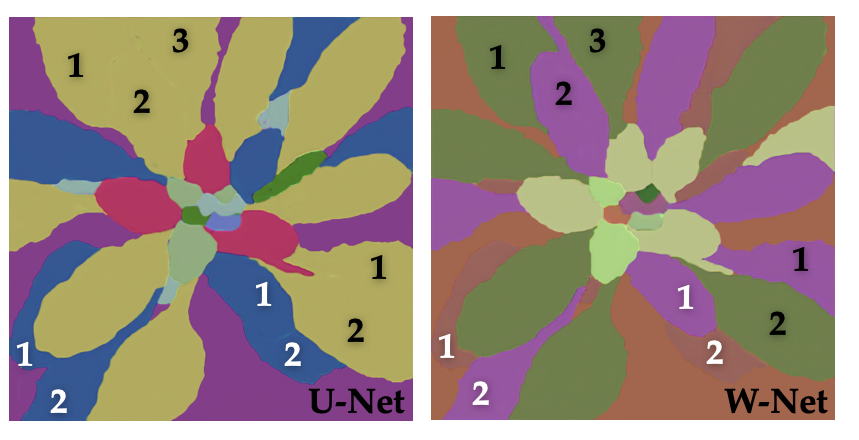}
	\end{subfigure}
	\caption{Learned Embeddings with U-Net and W-Net. Numbered leaves are treated as one object by U-Net, while they are successfully separated in the embedding space learned with W-Net.}
	\label{wnet}
\end{figure}

\subsection{U-Net vs. W-Net}
Firstly, we illustrate the performance improvement from U-Net with two heads to the proposed W-Net. In Fig.~\ref{wnet}, two representative cases are demonstrated, where the U-Net fails to separate closely located leaves. In contrast, the W-Net has successfully distinguished the numbered leaves in Fig.~\ref{wnet}.

Quantitatively, W-Net surpasses U-Net on overall mSBD by approximately 8\% from 0.794 to 0.879 with the best set-ups for W-Net, as shown in Table~\ref{tb:con}. Under different settings of embedding dimensions (Fig.~\ref{fig:dim_res}) and loss weights (Fig.~\ref{fig:inter_res}), the performance gap between U-Net and W-Net can be continuously observed and remain about 8\%.

\begin{table}
\centering
\begin{minipage}{.4\linewidth}
\caption{Comparison of Different Types of Concatenative Layers. Denotation: dfeat.16+efeat.16 = concatenated distance features of 16 dim and embedding features of 16 dim. Others can be analogously educed.}
\label{tb:con}
\centering
\begin{tabular}{ ccc}
\noalign{\smallskip}\toprule
Concatenative & \multirow{2}{*}{Net}& \multirow{2}{*}{mSBD}\\
Layer &&\\\midrule\noalign{\smallskip}
 none (baseline) & U-Net & .794\\
 coordinate &U-Net& .798\\
 distmap &W-Net& .824\\
 dfeat.8 &W-Net& .864\\
 dfeat.32 &W-Net& \textbf{.879}\\
 efeat.32 &W-Net& .847\\
dfeat.16+efeat.16 &W-Net & .873\\
\bottomrule
\end{tabular}
\end{minipage}
\qquad
\begin{minipage}{.52\linewidth}
\caption{Comparison of Local/Global Constraints, Network and Clustering. Denotations: Local = local constraints, otherwise global; 64d = 64 dim embeddings, otherwise 8 dim; AC = Angular clustering; MWS = Mutex Watershed~\cite{wolf2018mutex}.}
\label{tb:loss}
\centering
\begin{tabular}{cccc}
\noalign{\smallskip}\toprule
Local & Net & Clustering & mSBD\\
\midrule\noalign{\smallskip}
\(\cmark\)& W-Net & AC & \textbf{.879} \\
\(\cmark\)& W-Net 64d & AC & .854\\
 & W-Net & AC & .835 \\
& W-Net 64d & AC & .823\\\midrule
\(\cmark\)& U-Net & MWS & .719\\
\(\cmark\)& W-Net & MWS & .771 \\
\(\cmark\)& U-Net & MeanShift & .679\\
\(\cmark\)& W-Net & MeanShift & .733 \\
\(\cmark\)& U-Net & HDBSCAN & .631\\
\(\cmark\)& W-Net & HDBSCAN & .681 \\
\bottomrule
\end{tabular}
\end{minipage}
\end{table}

\subsection{Concatenative Layer}
We compare the effects of different types of concatenative layer. Firstly, the distmap (1-dimensional) can be directly forwarded. Alternatively, the distance regression features instead of the distmap can be utilized. Before concatenation, we convert the 32-channel \textit{D-feat.} into 8 and 32 dimensions (denoted as \textit{dfeat.8} and \textit{dfeat.32} in Table~\ref{tb:con}) through a single convolutional layer. 

Meanwhile, the case of using embedding loss as the intermediate supervision (\textit{efeat.32}) has also been tested. Specifically, the embedding features from the first U-Net are concatenated with the images as the inputs of the second embedding module. Furthermore, the concatenated distance regression features and embedding features (\textit{dfeat.16+efeat.16}) are also investigated.
At last, augmenting the input image with coordinates is tested. As proposed in~\cite{Novotny18b}, constructing dense object-aware pixel embeddings cannot be easily achieved using convolutions and the situation can be improved by incorporating information about the pixel location. In this work, we augment the input image with two coordinate channels for the normalized x- and y-coordinates, respectively.

Experimental results are summarized in Table~\ref{tb:con}. First of all, forwarding distmaps is not as effective as forwarding feature maps, including the distance regression features and the embedding features. The embedding features (\textit{efeat.32}) can also boost the performance, but not as significantly as the distance regression features. This is verified by the fact that \textit{efeat.32} is worse than \textit{dfeat.32} and the mixed feature map \textit{dfeat.16+efeat.16}. For the distance regression feature itself, higher dimensions of 32 are preferred. Finally, augmenting images with coordinates does not show apparent differences in our experiments. The effects could be further studied. For example, augmenting each intermediate feature map with coordinates is also worth being investigated. 

\subsection{Local vs. Global Constraints}
\label{exp:constraint}
Local constraints make it possible to exploit lower-dimensional embedding space more efficiently, as in this case, different labels only have to be distributed to the neighboring objects. In contrast, the global constraints have to thoroughly give each single object in the images a different label, which requires larger receptive fields and more redundant embedding space. The combination of local constraints and cosine embeddings utilizes the embedding space further comprehensively, as the push force imposed by loss expects orthogonal embedding clusters for neighboring instances.

\begin{figure}
	\centering	
	\begin{subfigure}{0.23\textwidth}
		\centering
    	\includegraphics[width=\textwidth]{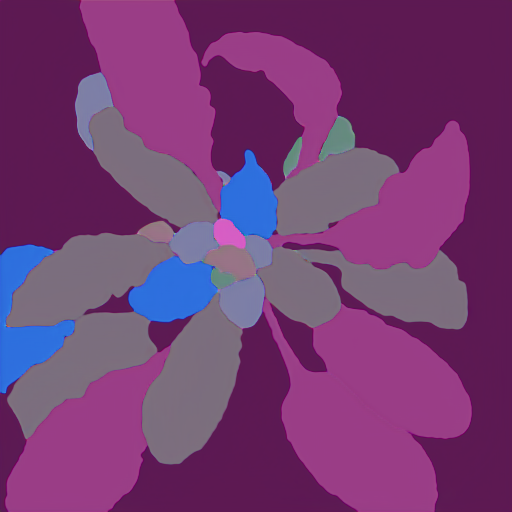}
    	\caption{local 8}
    	\label{fig:dim8}	
	\end{subfigure}
	\begin{subfigure}{0.23\textwidth}
		\centering
    	\includegraphics[width=\textwidth]{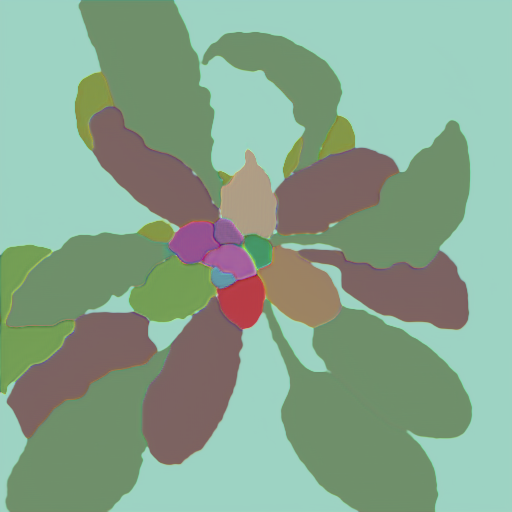}
    	\caption{local 64}
    	\label{fig:dim64}	
	\end{subfigure}
	\begin{subfigure}{0.23\textwidth}
		\centering
    	\includegraphics[width=\textwidth]{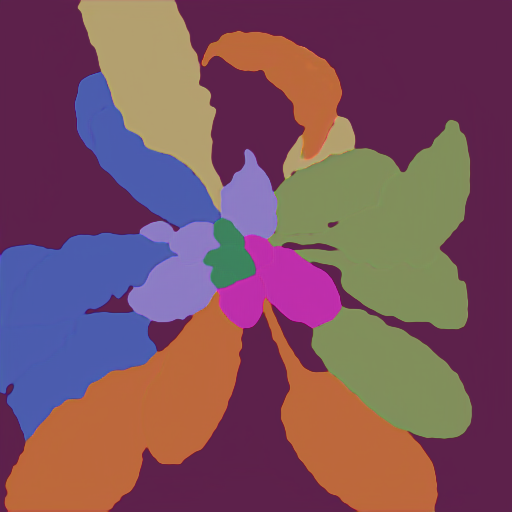}
    	\caption{global 8}
    	\label{fig:gdim8}	
	\end{subfigure}
	\begin{subfigure}{0.23\textwidth}
		\centering
    	\includegraphics[width=\textwidth]{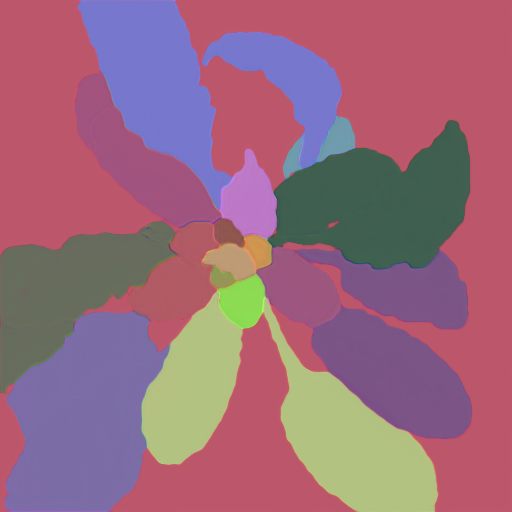}
    	\caption{global 64}
    	\label{fig:gdim64}	
	\end{subfigure}
	\caption{Learned Embeddings for Combined Cases of Local/Global Constraints and 8/64-dimensional Embeddings. (a,b) vs. (c,d): Local constraints ensure the effective utilization of embedding space, as same embeddings appear alternately for non-adjacent objects. (a) vs. (b): Higher-dimensional embeddings are redundant in the local constraint case. (c) vs. (d): Lower-dimensional embeddings with global constraints are not sufficient to distinguish all objects. This problem is slightly mitigated via higher-dimensional embeddings, still not as effective as local constraints.}
	\label{gdim}
\end{figure}

This is confirmed qualitatively by examples showcased in Fig.~\ref{gdim}. In Fig.~\ref{fig:gdim8}, 8-dimensional embeddings are trained with global constraints. Not surprisingly, there are exactly 8 colors in the image, indicating 8 orthogonal clusters in the embedding space. Apparently, the global constraint will fail when the embedding dimensions are fewer than the number of objects. In contrast, the local constraints (Fig.~\ref{fig:dim8} -~\ref{fig:dim64}) can distribute labels alternately between objects, with the same labels appearing multiple times for non-adjacent objects. This makes it possible to utilize a lower-dimensional embedding space. Quantitatively, the W-Net trained with local constraints surpasses the one trained with global constraints by more than 4\% on overall mSBD, as listed in Table~\ref{tb:loss}.

Intuitively, a higher-dimensional embedding space is able to provide a higher degree of freedom, i.e. we could simply use higher-dimensional embeddings to alleviate the problem of global constraints. At least the embedding vector does not have to be restricted to low dimensions. However, from the results in Fig.~\ref{fig:dim_res}, we find that higher-dimensional embeddings produce worse results. This makes the capability of using lower-dimensional embedding space particularly important. 

\subsection{Dimensions of Embeddings}\label{ssec:dim}
\begin{figure}
	\centering	
	\begin{subfigure}{0.48\textwidth}
		\centering
    	\includegraphics[width=\textwidth]{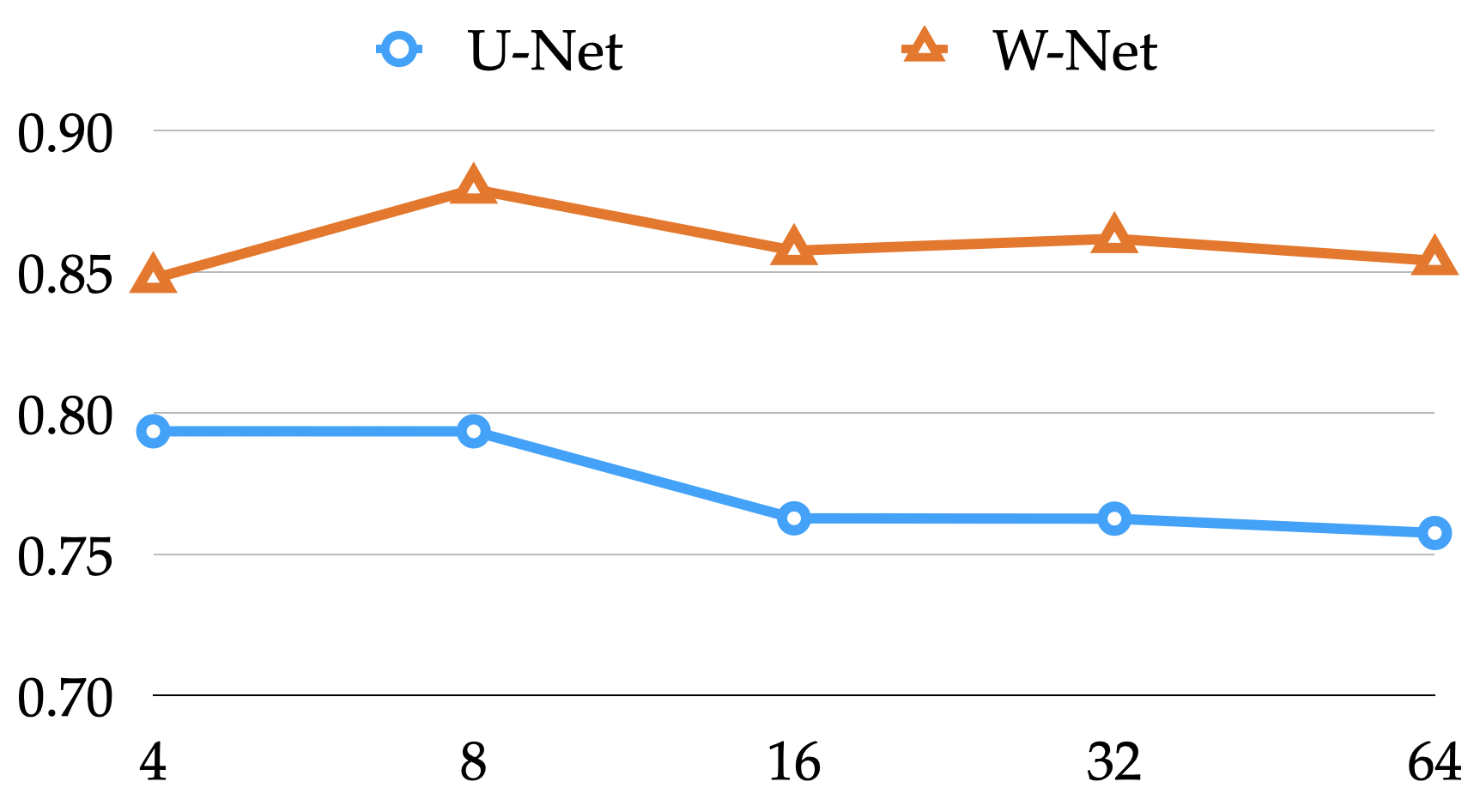}
    	\caption{Dimensions of Embeddings}
    	\label{fig:dim_res}	
	\end{subfigure}
	\hfill
	\begin{subfigure}{0.48\textwidth}
		\centering
    	\includegraphics[width=\textwidth]{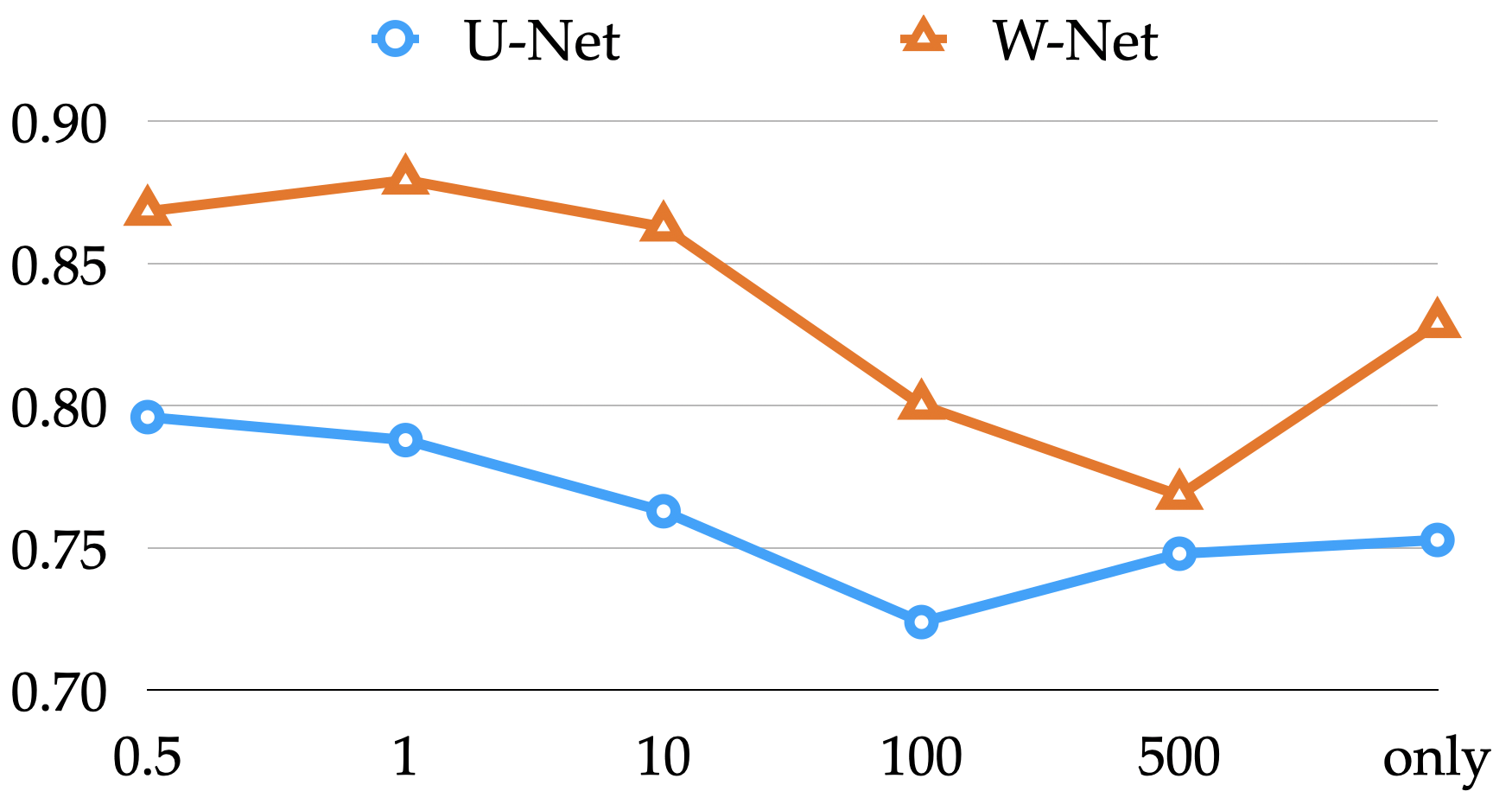}
    	\caption{Loss Weights \(\lambda\)}
    	\label{fig:inter_res}	
	\end{subfigure}
	\caption{mSBD w.r.t. Dimensions of Embeddings and Loss Weight \(\lambda\) in \(\mathcal{L}_{emb}=\lambda \cdot \mathcal{L}_{inter} + \mathcal{L}_{intra}\) using U-Net and W-Net. In (b), \textit{only} denotes \(\mathcal{L}_{emb} =\mathcal{L}_{inter}\). W-Net surpasses U-Net generally. Best overall performance of W-Net can be obtained with 8-dimensional embeddings and \(\lambda=1\).}
	\label{plot:1}
\end{figure}

As discussed previously, the local constraints make the use of lower-dimensional embedding possible. It is thus worth investigating the influence of different embedding dimensions on the overall performance. The mSBD scores of both U-Net and W-Net for \{4, 8, 16, 32, 64\}-dimensional embeddings are plotted in Fig.~\ref{fig:dim_res}. For 32 and 64 dimensions, the batch size is set to 2, instead of 4 as in other cases, to fit the memory of a single GPU. 

Our experiments show that the 8-dimensional embedding brings in the best result. First of all, merely 4 dimensions are incompetent to separate all adjacent objects, since it is common that one object has more than 4 neighbors. Although higher dimensions may not bring in more labels under local constraints, comparing Fig.~\ref{fig:dim8} to~\ref{fig:dim64}, increasing the embedding dimensions should not degrade the performance hypothetically. However, the mSBD score decreases slightly as the dimensions increase. Therefore we believe, under the premise that the dimensions are sufficient for all objects to fulfill the local constraints, higher-dimensional embedding space is more difficult to train.

\subsection{Loss Weights}
During the experiments, we find that the values of between-instance loss term \(\mathcal{L}_{inter}\) are approximately 10 times greater than the values of within-instance loss term \(\mathcal{L}_{intra}\). This is consistent with the fact that pixel embeddings of the same object converge tightly, but adjacent objects are not correctly segmented occasionally. The larger weighting factor \(\lambda\) of between-instance loss term \(\mathcal{L}_{inter}\) might be helpful to emphasize the significance of it by amplification of its gradient. We set \(\lambda\) as \{0.5, 1, 10, 100, 500\}, and moreover, we omit the within-instance loss, denoted as \textit{only} in Fig.~\ref{fig:inter_res}. The experiments are preformed for both U-Net and W-Net under identical main set-ups: 32-dimensional distance features as concatenative layer, local constraints and 8-dimensional embeddings. 

\begin{figure}
	\centering	
	\begin{subfigure}{0.23\textwidth}
		\centering
    	\includegraphics[width=\textwidth]{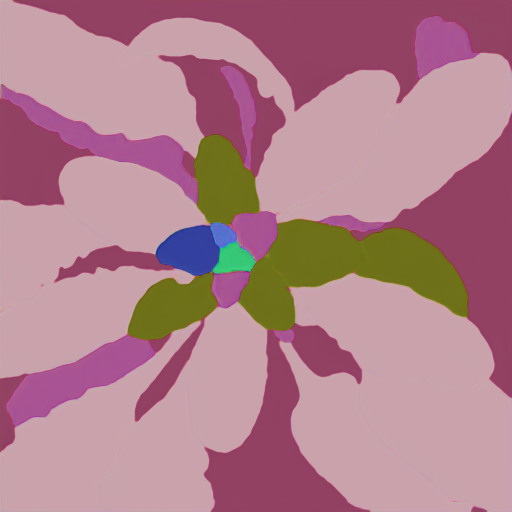}
    	\caption{\(\lambda=0.5\)}
    	\label{fig:half}	
	\end{subfigure}
	\begin{subfigure}{0.23\textwidth}
		\centering
    	\includegraphics[width=\textwidth]{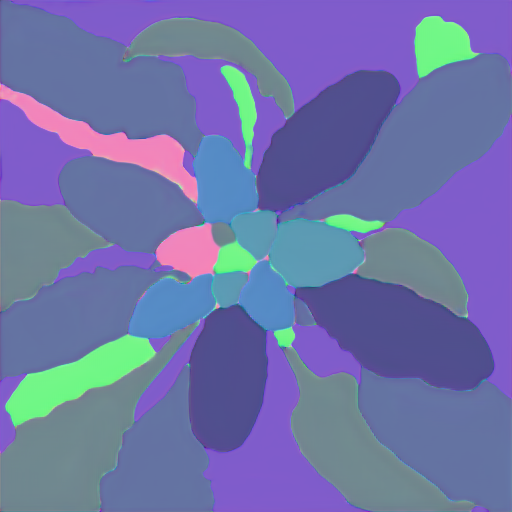}
    	\caption{\(\lambda=10\)}
    	\label{fig:ten}	
	\end{subfigure}
	\begin{subfigure}{0.23\textwidth}
		\centering
    	\includegraphics[width=\textwidth]{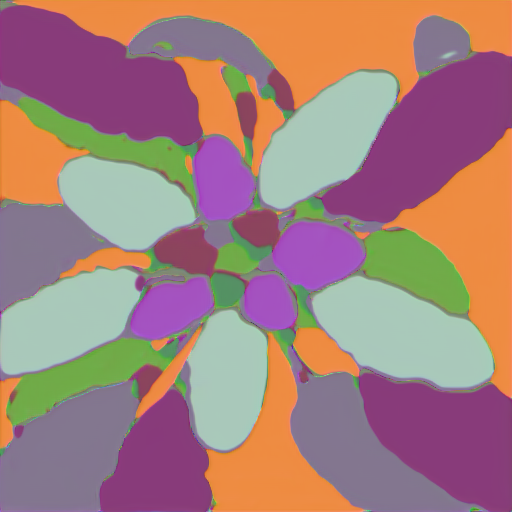}
    	\caption{\(\lambda=100\)}
    	\label{fig:hund}	
	\end{subfigure}
	\begin{subfigure}{0.23\textwidth}
		\centering
    	\includegraphics[width=\textwidth]{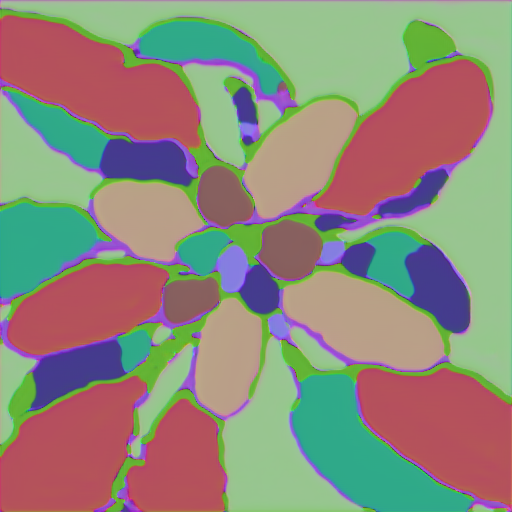}
    	\caption{only \(\mathcal{L}_{inter}\)}
    	\label{fig:only}	
	\end{subfigure}
	\caption{Learned Embeddings with Different Weights \(\lambda\) as in \(\mathcal{L}_{emb} = \lambda \cdot\mathcal{L}_{inter} + \mathcal{L}_{intra}\). With ascending \(\lambda\), overall segmentation performance becomes worse (Fig.~\ref{plot:1}) with the decreased consistency of embeddings in the same object. It is worth noting that training with just the between-instance loss term can also to some extent form clusters in the embedding space.}
	\label{weights}
\end{figure}

From the experiments, we find that larger weighting factor of the between-instance loss term does not further help to encourage the network to separate the confused objects when \(\lambda\) is larger than 1, but reduces the consistency of embeddings in the same object. Fig.\ref{weights} showcases the trade-off between the discrimination of adjacent objects (larger \(\lambda\)) and the consistency of individual object (smaller \(\lambda\)). The experiments show that \(\lambda=1\) brings in best overall performance, as shown in Fig.~\ref{fig:inter_res}. Besides, one surprising conclusion is that training the network with just the between-instance loss term can also, to some extent, form clusters in the embedding space (Fig.~\ref{fig:only}).

\subsection{Clustering}
Apart from the default angular clustering used along through the experiments, other three clustering techniques have been tested based on the predicted embeddings of the best results: Mutex Watershed~\cite{wolf2018mutex}, Mean Shift~\cite{fukunaga1975estimation} and HDBSCAN~\cite{campello2015hierarchical}. On the one hand, this provides a reference for the performance of different clustering methods on the embeddings trained with cosine similarity based loss. On the other hand, it can also indirectly reflect the quality of embeddings generated by U-Net and W-Net. Results are shown in Table~\ref{tb:loss}. 

In conclusion, the angular clustering has advantages in terms of performance and speed. Nevertheless, it should be noted that this method is only applicable to the case, where seeds are available and clusters are orthogonal in the embedding space. Additionally, all clustering approaches produce better results with embeddings predicted from the W-Net, which again confirms the improvement of our proposed method.

\begin{table}
\caption{Comparison of Results. Abbreviations: Aug. = Data augmentation; Emb. = Metric of embedding similarity; Fg. = Ground truth foreground masks are used; syn = Synthetic images are used for training; HG = Stacked Hourglass network; Lb. = Results shown in the leaderboard of CodaLab.}
\label{tb:ev}
\centering
\begin{tabular}{lcc ccc c ccc}
\noalign{\smallskip}\toprule
\multirow{2}{*}{Method} & \multirow{2}{*}{Backbone} & \multirow{2}{*}{Train} & \multirow{2}{*}{Aug.} &\multirow{2}{*}{Emb.}  & \multirow{2}{*}{Fg.} & \multirow{2}{*}{Lb.} & \multicolumn{3}{c}{mSBD} \\\cmidrule(lr){8-10}
 &&& &&& & A1 & A1-3 & A1-5\\
\midrule
 IPK~\cite{ipk,Leaf} & - & A1-3 & &&\(\cmark\)          & & .791 & .782 & - \\
 Nottingham~\cite{Leaf} &-&A1-3&  &&      \(\cmark\)     && .710 & .686 & - \\
 MSU~\cite{Leaf,msu} &-&A1-3& &&                 \(\cmark\)&      & .785 & .780 & - \\
 Wageningen~\cite{Leaf} &-&A1-3& &&             \(\cmark\)          & & .773 & .769 & - \\
 \midrule
 MRCNN~\cite{chen2019instance,he2017mask} & ResNet &A1-3& &&&              &  - & .797 & -\\
 Stardist~\cite{chen2019instance,schmidt2018}  & U-Net &A1-3& &&&                   & - & .802 &- \\
 IS-RA~\cite{ren2017end} & FCN & A1 & &&&                     &  .849 & - & - \\
 Ward~\cite{ward2018deep} & ResNet  & A1-4+syn & \(\cmark\)  &&&                     &  .900 & .740 & .810 \\
 UPGen~\cite{ward2020scalable} & ResNet & A1-4+syn & \(\cmark\) &   &&                     & .890  & \textbf{.877} & .874 \\
\midrule
DiscLoss~\cite{de2017semantic} & ResNet & A1 & \(\cmark\) & euc & \(\cmark\) & & .842  & - & - \\
CE-RH~\cite{payer2018instance} & HG & A1 & \(\cmark\) &cos&  & & .845 & - & -\\
E-LC~\cite{chen2019instance} & U-Net & A1-3 & &cos&&  & - & .831 & .823 \\
W-Net (ours) & U-Net & A1-4 & & cos & \(\cmark\)& \(\cmark\) & \textbf{.919} &  .870 & \textbf{.879} \\
\bottomrule
\end{tabular}
\end{table}

\subsection{Comparison against State-of-the-Art}
Comparison of state-of-the-art methods on the CVPPP LSC dataset is quantitatively shown in Table~\ref{tb:ev}. It is clear that the learning based methods (denoted with backbones) can achieve better results than the first four classical methods. The last four methods are based on pixel embedding learning. Roughly speaking, they bring in promising results. Our overall result mSBD for A1-5 outperforms all others. In the leaderboard, our overall result is at the 1. position by paper submission. Furthermore, the average of mSBD scores for Arabidopsis images (A1, A2, A4) outperforms the second best results from three different users, respectively, by over 3\%, namely 0.883 to 0.917. Due to the extremely imbalanced training images on Arabidopsis (783 images) and Tobacco (27 images), our result on testing set A3 are not as good as others, with mSBD of 0.77. Compared to this, the current 1. place mSBD of A3 in the leaderboard reaches 0.89. It implies that the sufficient number of training images is critical in our proposed method. We leave this room for improvement in the future. One thing worth mentioning is that the authors tend to not submit their results to the leaderboard of CodaLab, which makes the consistent comparison and review rather difficult. 
 
 \subsection{Application to Human U2OS Cells}
\begin{figure}[t]
	\centering	
	\begin{subfigure}[t]{0.3\textwidth}
		\centering
    	\includegraphics[width=\textwidth]{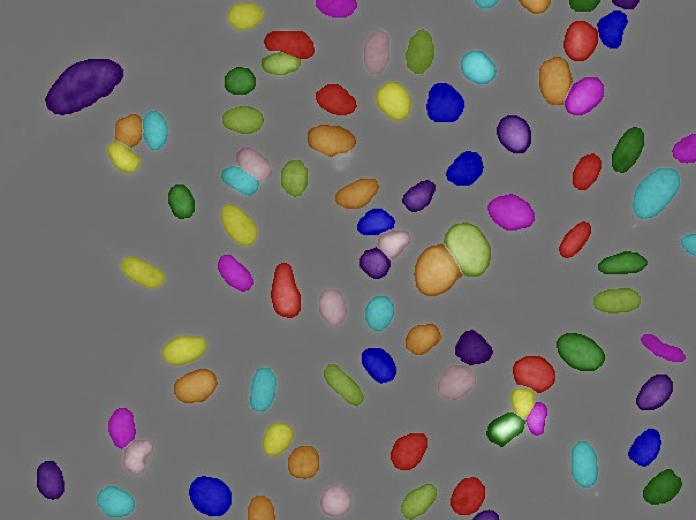}
	\end{subfigure}
		\begin{subfigure}[t]{0.3\textwidth}
		\centering
    	\includegraphics[width=\textwidth]{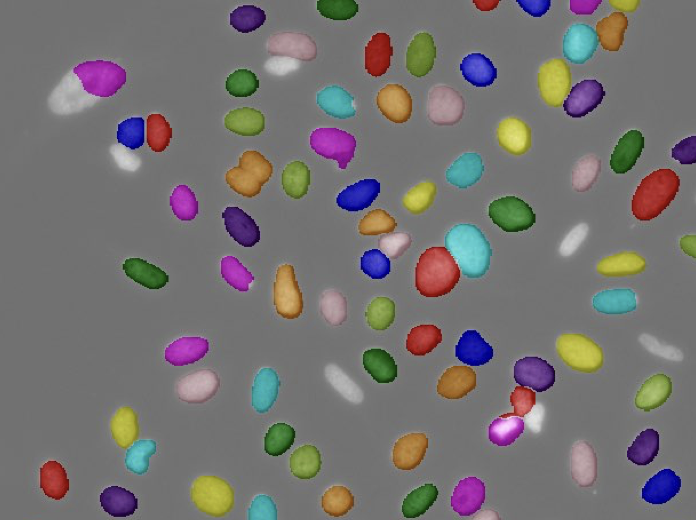}
	\end{subfigure}
		\begin{subfigure}[t]{0.3\textwidth}
		\centering
    	\includegraphics[width=\textwidth]{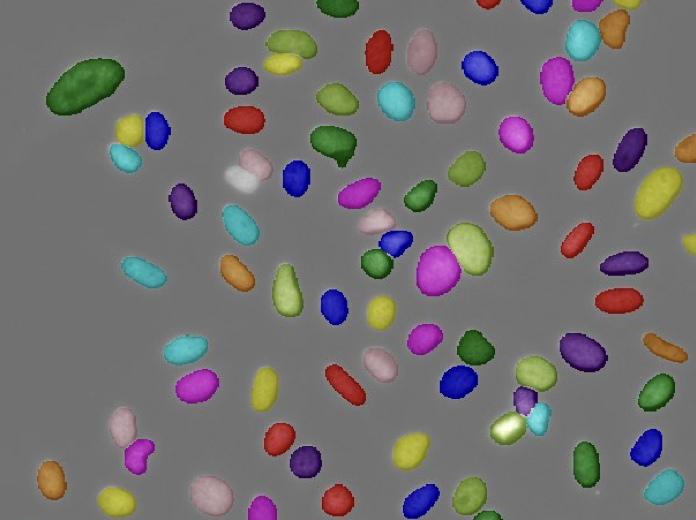}
	\end{subfigure}\\\smallskip
	\begin{subfigure}[t]{0.3\textwidth}
		\centering
    	\includegraphics[width=\textwidth]{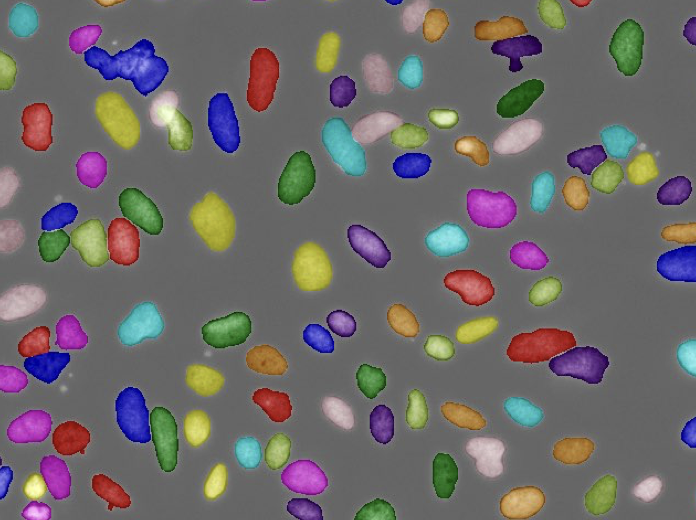}
	\end{subfigure}
		\begin{subfigure}[t]{0.3\textwidth}
		\centering
    	\includegraphics[width=\textwidth]{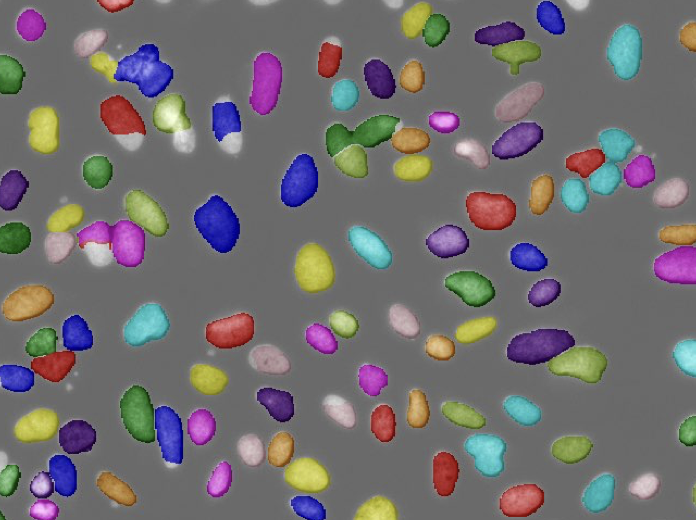}
	\end{subfigure}
		\begin{subfigure}[t]{0.3\textwidth}
		\centering
    	\includegraphics[width=\textwidth]{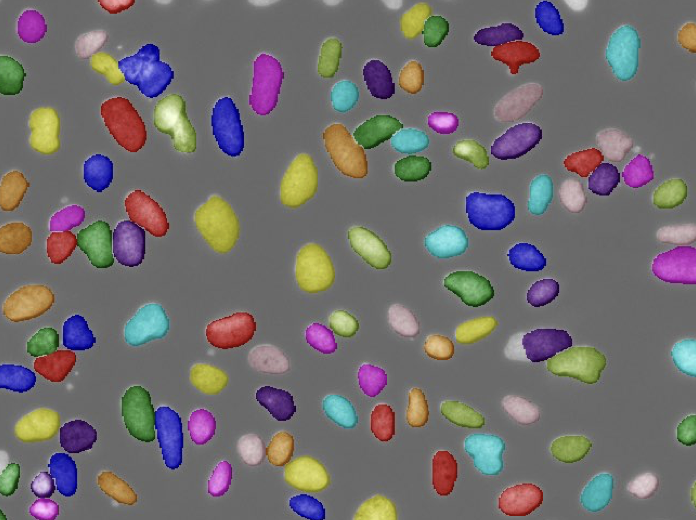}
	\end{subfigure}
	\caption{Cell Segmentation Results of the BBBC006v1 Data: Ground Truth (left), U-Net (middle) and W-Net (right); Each row illustrates one example. The improvement from U-Net to W-Net is salient. The mSBD and mAP scores have increased from 0.896 to 0.915 and from 0.577 to 0.664, respectively.}
	\label{fg:cell}
\end{figure}

Our method has also been tested on the image set BBBC006v1 of human U2OS cells from the Broad Bioimage Benchmark Collection~\cite{ljosa2012annotated}. Totally 754 images are randomly separated into two equally distributed training and testing set with 377 images respectively. Other set-ups are identical to previously introduced ones. We use U-Net and W-Net with distance concatenative layer to show results in mSBD and mean Average Precision with IoU=\{0.5, 0.55, 0.6, ..., 0.9\} (mAP). The mSBD has increased from 0.896 to 0.915 and the mAP from 0.577 to 0.664. We showcase two examples of final labels in Fig.~\ref{fg:cell}. As reported in~\cite{chen2019instance}, some embeddings around boundaries might be incomplete, which leads to incomplete segmentations. This problem has been mainly solved, as showcased in Fig.~\ref{fg:cell}.

\section{Conclusion}
In this work we propose a novel W-Net, which forwards the distance regression features learned by the first-stage U-Net to the subsequent embedding learning module. The intermediate distance regression supervision effectively promotes the accuracy of learned pixel embedding space, with the mSBD score on the CVPPP LSC dataset increased by more than 8\% compared to the identical set-up without supervision of distance regression features. We have also conducted a number of experiments to investigate the characteristics of the pixel embedding learning with the cosine similarity based loss, involving the embedding dimensions, the weighting factor of the within-instance loss term and the between-instance loss term. We are looking forward to applying this method to more datasets in the future.
\subsubsection{Acknowledgments} This work was supported by the German Research Foundation (DFG)
Research Training Group 2416 MultiSenses-MultiScales.

\clearpage
%
%
\bibliographystyle{splncs04}
\bibliography{bib}
\end{document}